\documentclass[a4paper,conference]{IEEEtran}
\ifCLASSINFOpdf
\else
\fi

\usepackage[utf8x]{inputenc} 
\usepackage{times}
\usepackage{epsfig}
\usepackage{graphicx}
\usepackage{amsmath}
\usepackage{amssymb}
\usepackage{multirow}
\usepackage{multicol}

\usepackage{algorithm}
\usepackage{algorithmicx}
\usepackage{algpseudocode}
\usepackage{amsmath}
\usepackage{float} 
\usepackage{subfigure}
\usepackage{ulem}
\usepackage{mathtools}
\usepackage{multirow}
\usepackage{graphicx}
\usepackage{paralist}

\usepackage{xcolor}
\usepackage[colorlinks = true,
            linkcolor = blue]{hyperref}
\graphicspath{{figs/}}
\begin{document}
%
\title{Predicting Team Performance with Spatial Temporal Graph Convolutional Networks
}

\author{\IEEEauthorblockN{Shengnan Hu}
\IEEEauthorblockA{Department of Computer Science\\
University of Central Florida\\ 
Orlando, FL, USA\\
Email: shengnanhu@knights.ucf.edu}
\and
\IEEEauthorblockN{Gita Sukthankar}
\IEEEauthorblockA{Department of Computer Science\\
University of Central Florida\\ 
Orlando, FL, USA\\
Email: gitars@eecs.ucf.edu}}


%


\maketitle

\begin{abstract}
This paper presents a new approach for predicting team performance from the behavioral traces of a set of agents.  This spatiotemporal forecasting problem is very relevant to sports analytics challenges such as coaching and opponent modeling.  We demonstrate that our proposed model, Spatial Temporal Graph Convolutional Networks (ST-GCN), outperforms other classification techniques at predicting game score from a short segment of player movement and game features. Our proposed architecture uses a graph convolutional network to capture the spatial relationships between team members and Gated Recurrent Units to analyze dynamic motion information.  An ablative evaluation was performed to demonstrate the contributions of different aspects of our architecture. 
\end{abstract}


%
\IEEEpeerreviewmaketitle

\section{Introduction}
Spatiotemporal forecasting is often applied to modeling and predicting large scale patterns caused by traffic, weather, and disease spread~\cite{li2018diffusion}. This paper demonstrates that it is possible to leverage spatiotemporal traces generated from a small number of moving entities by combining multiple types of features.  We propose a new neural network architecture, Spatial Temporal Graph Convolutional Networks (ST-GCN), that uses a weighted graph to represent the spatial relationships between agents and recurrent neural network layers to capture temporal movement patterns.  

Our research is devoted to the problem of analyzing, predicting, and modeling human teamwork.  Physical team tasks can be analyzed using the following types of cues: 1) spatial relationships between team members, 2) temporal dependencies of actions, 3) coordination requirements between agents who are working towards a joint goal~\cite{Sukthankar-ACMTISTActivityRecognition2011}.  Human teams are also common in virtual environments, including massively multiplayer online games~\cite{teamperformance} and military training simulations~\cite{Sukthankar-MOO2005}.    Our proposed architecture, ST-GCN, encodes spatial and coordination information using a Graph Convolutional Network (GCN) to aggregate feature vectors from different team members, and Gated Recurrent Units (GRUs) to encode the temporal structure of team plans.

Although there exists a rich literature on multi-agent activity recognition~\cite{Sukthankar-ACMTISTActivityRecognition2011,beetz2005,chen2014play}, our aim is to predict the performance of the team rather than to recognize behavior sequences.   This is an important problem in sports analytics, a specialized form of data mining which uses player and team statistics to extract actionable insights for coaching, training, and recruitment~\cite{sportsanalytics,sportsperformance}.  We seek to monitor team performance from observing a combination of spatiotemporal behavior traces and game-based features in order to create an agent who can offer coaching advice when the team is exhibiting poor performance.

There is previous work on analyzing games with defined field structure and goal regions, including American football~\cite{chen2014play}, soccer~\cite{beetz2005}, and basketball~\cite{sanguesa2017identifying}; however this paper tackles the problem of modeling player performance in domains with more complex terrain features such as military simulations and massively multiplayer online games.  A second challenge is handling teamwork domains when there is only a modest amount of data; due to the expense of data collection, much of the research on team cognition has been performed on small datasets with fewer than thirty teams~\cite{salas2004team}.  Our proposed method achieves a lower sample complexity by intelligently fusing a rich set of features extracted from player field of view using our Graph Convolutional Network. 

Graph neural networks are used to express domains where there are complex interdependencies between entities~\cite{gnnsurvey}; GCNs aggregate feature information from neighboring nodes through the inclusion of an adjacency matrix in the forward pass operation.  Like standard convolutional networks, the GCN also achieves parameter reduction through weight sharing.  Unlike a standard convolutional network, node neighbors have no implicit ordering and can be variable in number.  

We present results on predicting the team performance of teams executing a simulated search and rescue task within Minecraft.  Minecraft is a 3D sandbox video game that can support many modes of game play and is the testbed for several AI challenge problems~\cite{johnson2016malmo}, including the MineRL competition~\cite{minerl}.  Our proposed architecture, ST-GCN, is used to predict the score of the team at rescuing victims over a short time segment, and we performed an ablative evaluation to demonstrate the utility of each addition to our neural architecture.   The next section provides an overview of related work in the area.





\begin{figure*}
\centering
\includegraphics[width=0.8\textwidth]{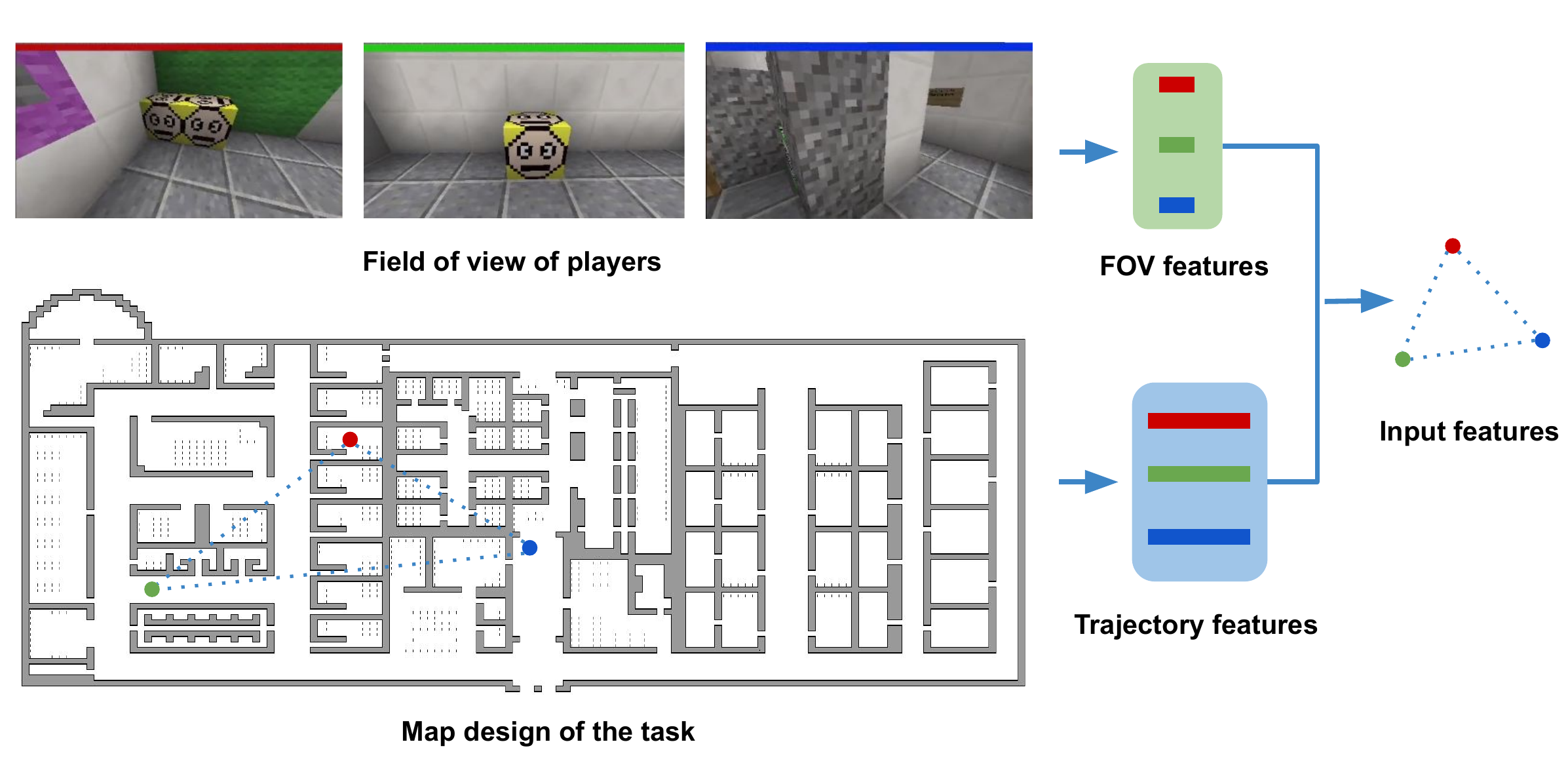}
\caption{Our proposed architecture fuses features visible in the players' field of view with trajectory features.}
\label{fig:feature}
\end{figure*}

\section{Related Work}
Wu et al.~\cite{gnnsurvey} organize graph neural networks into a taxonomy of 1) recurrent graph neural networks, 2) convolutional graph neural networks, 3) graph autoencoders, and 4) spatial-temporal graph neural networks.  Our proposed technique is an example of a spatial-temporal graph neural network (STGNN) which learns patterns from spatio-temporal input.  This style of network has been very successful at spatio-temporal forecasting~\cite{tgcn,li2018diffusion}, particularly in the area of traffic prediction where time series models such as ARIMA are unable to encode the spatial dependencies resulting from road network connectivity. 

Like our proposed architecture, the T-GCN (Temporal Graph Convolution Network) technique~\cite{tgcn} for traffic prediction incorporates both GCN and GRU layers. However, T-GCN applies a set of GCN and GRU modules to directly predict the traffic at time $t+1$. Our method uses GCN and GRU layers as a feature extractor to aggregate node features; it outputs feature vectors at every timestamp.  Prediction is then done using a global average pooling layer to concatenate features across multiple time steps to perform the prediction.  Also in traffic forecasting the adjacency matrix is based on the road network and is fixed; in our dataset, the adjacency matrix is extracted from player position and keeps changing as the participants move. Another approach is to create a graph from a combination of spatial and temporal relationships and then to apply the GCN to the graph.  For instance, the space-time region graph model leverages both similarity and spatial-temporal relations in the graph for human action recognition~\cite{wang2018videos}. In this paper we benchmark our work against DCRNN (Diffusion Convolution Recurrent Neural Network)~\cite{li2018diffusion}, a top performing traffic prediction method.  The diffusion process is modeled as a random walk on the graph with a restart probability. DCRNN combines diffusion convolutions with GRUs and learns a set of trainable weights assuming a K-step truncation of the diffusion process. 

Teamwork analysis and performance prediction have also been studied within the multi-agent systems community.  STABR (Simultaneous Team Assignment and Behavior Recognition)~\cite{Sukthankar-ACMTISTActivityRecognition2011} was developed to analyze movement traces generated by dynamic teams who can perform multiple behaviors in parallel; it uses RANSAC to fit static spatial models to team groupings in order to recognize behaviors.  Like STABR, ST-GCN can handle cases in which players stop collaborating with one another and independently take point scoring actions. Weighted synergy graphs~\cite{synergygraph} were developed as mechanism to predict how well an ad-hoc team will perform at a task. In contrast, the ST-GCN graph can be created even when coordination constraints are not explicitly known since the adjacency matrix is based on player position.  The next section provides background on our application domain.


\section{Background}
The ultimate aim of our research is to be able to predict the performance of human teams to facilitate the development of proactive assistant agents who help teams by anticipating and detecting teamwork failures. Our research was conducted using a variant of the Minecraft testbed created to evaluate human-machine teamwork~\cite{asisttestbed}.  It is based on Malmo~\cite{johnson2016malmo}, a general AI experimentation platform, designed to support the development of Artificial General Intelligence agents.   Minecraft players perform many types of collaborative activities including group construction projects, team combat, and puzzle solving.  Server side monitoring tools, such as Heapcraft~\cite{muller2015quantifying}, have been created for studying human collaboration in Minecraft based on player proximity, item sharing, and communication. 

We trained our ST-GCN model on data \cite{ASU/BZUZDE_2022} from human teams executing a search and rescue (SAR) operation.  SAR teams consist of three players searching for victims in a building simulated within Minecraft; there are two scenarios (Mission A and B). Team members can serve different roles, including acting as medic (healing victims), searcher (finding victims), or engineer (destroying rubble that impedes the search).  Players can communicate either by audio on Zoom or through annotating the area with marker blocks to alert other players about victims that need to be rescued.  The task was specifically designed to test human teamwork; more points are gained by saving victims that require multiple rescuers.  The team objective is to gain as many points as possible by triaging the victims within the time limit of 15 minutes.  To predict the team performance, our model needs to effectively encode what the players see and do in the environment; since they must coordinate with each other to rescue heavily wounded victims, spatial proximity between players is important as well. 


\section{Approach}


\subsection{Spatial Graph Construction}
\label{sec:gcn}
Given a spatiotemporal behavior trace,  we denote it as $V = \left\{I_{1}, I_{2}, ..., I_{T}  \right\}$, where T is the number of windows sampled from the trace. For each window, we designate the feature extracted by our system as $X^{(t)}$. As shown in Fig. \ref{fig:feature}, the feature vector of each participant contains two parts: the field of view (FOV) features and trajectory features. We obtain the FOV features using CMU's PyGL-FoV software package which outputs the a list of the blocks that appear in the monitor port viewed by the participant.  For this task, we focus on the appearance of victims, and denote the FOV feature at time $t$ as $X^{(t)}_{fov} = k$ , where $k$ is the number of victims visible to the participant. The trajectory features are defined as  $X^{(t)}_{traj} = \left\{x, y, v, v_x, v_y \right\}$, where $(x, y)$ and $v$ represent the coordinates and the velocity of the participant at time $t$.  $v_x$ and $v_y$ are the velocities in the $x$ and $y$ directions. Thus, we have the input node feature $X^{(t)} = X^{(t)}_{fov} + X^{(t)}_{traj}$. 

By treating each player in the team as a node, we construct a weighted graph $G = (V, E, A^{(t)})$ to describe the topological structure of the team, This graph has the set of nodes $V = \{v_{1}, v_{2}, ..., v_{N}\}$, where $N$ denotes the number of nodes; for the SAR task $N = 3$. $E$ is the set of edges, and the edges $e_{ij} = (v_{i}, v_{j}) \in E$ are used to represent the relationships between participants.  Since all the three participants are assumed to be working together, we set all the edges in $E$ to 1, which means that all the participants are connected with one another. 

The adjacency matrix $A^{(t)} \in \mathbb{R}^{N*N}$ describes the edge weights of the graph at time $t$, which can also be considered as the connectivity information between the nodes. This is the key component for the GCN construction. In our model, the adjacency matrix  $A_{ij}^{(t)}$ is defined by the Euclidean distance $d_{ij}$ between participants $v_{i}$ and $v_{j}$,
\begin{equation}
 d(v_i, v_j) = \sqrt {\left( {x_i - x_j } \right)^2 + \left( {y_i - y_j } \right)^2 },
\end{equation}
where $(x_i, y_i)$ and $(x_j, y_j)$ are the coordinates of participant $v_i$ and $v_j$. Also, the following three constraints need to be considered when building the adjacency matrix: 
\begin{compactitem}
\item each element in $A$ needs to be non-negative with a value in the range of (0, 1)
\item the sum of each row of the adjacency matrix should be 1
\item for each element of $A$, the value of $A_{ij}$ should be inversely proportional to the Euclidean distance  between $v_{i}$ and $v_{j}$. 
\end{compactitem}
Thus, we perform normalization on each row of $A$ such that: 
\begin{equation}
A_{ij}^{(t)} = \frac{exp(-d(v_i, v_j))}{\sum_{i=1}^{N} exp(-d(v_i, v_j))}.
\end{equation}

Specifically, as the participants move away from each other, the connection between them becomes weaker, and the value of  $A_{ij}$ becomes smaller, and vice versa. Inspired by \cite{kipf2016semi}, we have $\widetilde{A} = A^{(t)} + I_{N}$, which represents a self-connected matrix, with $I_N$ defined as the identity matrix. This yields a normalized graph Laplacian matrix
\begin{equation}
\widehat{A} = \widetilde{D}^{-\frac{1}{2}}\widetilde{A}\widetilde{D}^{-\frac{1}{2}},
\end{equation}
where $\widetilde{D}$ is a diagonal matrix of node degrees and $\widetilde{D}_{ii} = \sum_{j}(A^{(t)}_{ij})$.

Finally, we adopt a 2-layer GCN model \cite{bruna2013spectral} with rectified linear units to learn the spatial features of the team:
\begin{equation}
\label{con:fxa}
f(X^{(t)}, A^{(t)}) = \sigma (\widehat{A}\mbox{ReLU}(\widehat{A}X^{(t)}\Theta_0)\Theta _1),
\end{equation}
where $\Theta_0$ and $\Theta_1$ are weight matrices with learnable parameters in the first and second graph layer.

\subsection{Temporal GCN Module} 
Although the GCN captures the spatial relationships between participants, it is hard for GCN to monitor the dynamic motion information of the players. Here, we leverage recurrent neural networks (RNNs) to facilitate the extraction of the temporal  information across multiple frames.  Due to their computing efficiency, we chose Gated Recurrent Units (GRU), which are simple yet powerful. Then, we use the GCN output feature as the input of GRU to model the temporal dependencies, and finally we have,

\begin{gather}
r^{t} = \sigma (\Theta _{r}[f(X^{(t)}, A^{(t)}), H^{(t-1)}] + b_{r}) \\
u^{t} = \sigma (\Theta _{u}[f(X^{(t)}, A^{(t)}), H^{(t-1)}] + b_{u}) \\
C^{(t)} = \tanh (\Theta _{c}[f(X^{(t)}, A^{(t)}), (r^{(t)} \odot H^{(t-1)})] + b_{c}) \\
H^{(t)} =  u^{(t)} \odot  H^{(t-1)} + (1-u^{(t)}) \odot  C^{(t)},
\end{gather}

where $X^{(t)}$ is the input player information at time $t$; $H^{(t-1)}$ denotes the hidden state at time $t - 1$; $H^{(t)}$ is the output at time $t$; $r^{(t)}$ is the reset gate at time $t$, which controls the degree to which past information is ignored; $u^{(t)}$ is the update gate at time $t$, which controls what information is included in the current status. $f(X^{(t)}, A^{(t)})$ is the graph convolution operation defined in Equation \ref{con:fxa}. $\Theta$ and $b$ are the weights and deviations in the training process, and $\odot$ denotes element-wise multiplication. The player information at time $t$ and the hidden state at time $t-1$ are combined by the GRU module to form the output vector at time $t$. Thus our proposed model not only leverages the players' current features, but also preserves historical information.

\begin{figure*}
\centering
\includegraphics[width=0.8\textwidth]{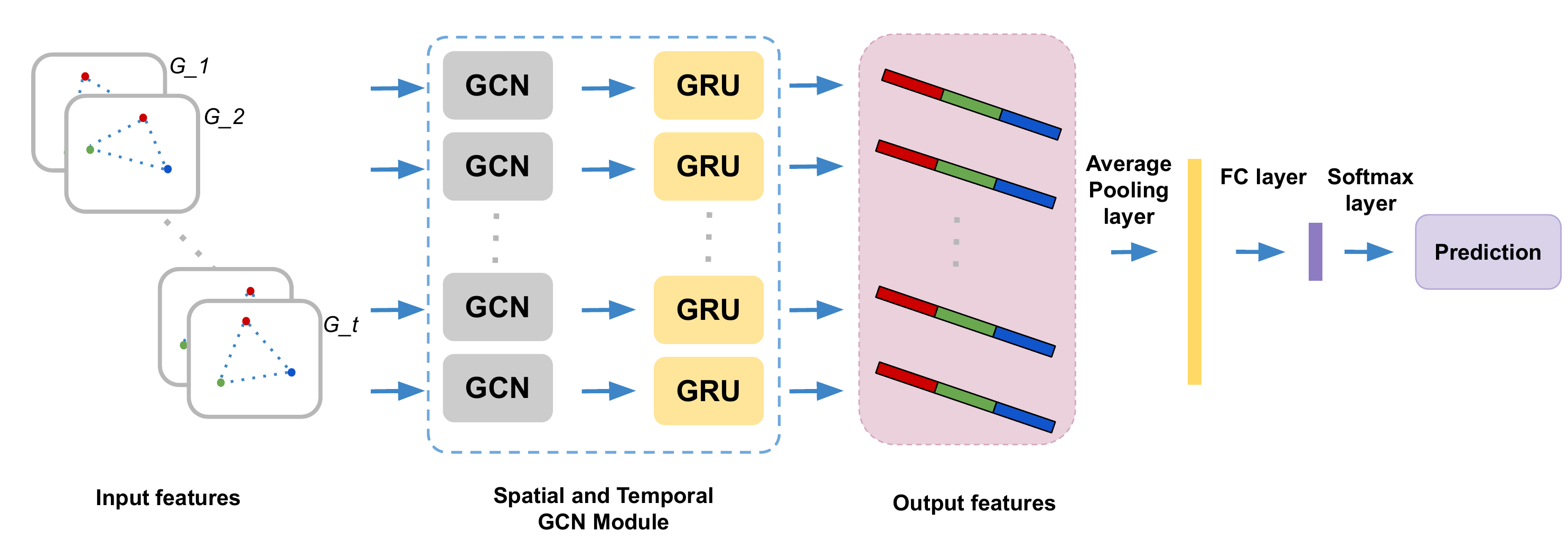}
\caption{The architecture of our proposed method, ST-GCN.  Input features are fed to the GCN using an adjacency matrix connecting team members by a value inversely proportional to their separation.  The output of the GCN is passed through a GRU.  The GRU outputs across a short temporal segment are combined using global average pooling.  The final result is passed through a fully connected layer and softmax to make the prediction of team performance.}
\label{fig:model}
\end{figure*}


\subsection{Loss Function}
The architecture for the proposed Spatial Temporal Graph Convolutional Networks (ST-GCN) model is shown in Fig. \ref{fig:model}. By applying both GCN and GRU modules, our proposed model can capture both the spatial dependence and temporal changes of the team. Given the final output $H^{(t)}$ of GRU at time $t$, we apply a dimensionality reduction operation and combine the features of all the nodes into a single vector. 

To concatenate the features from each window into a single compact descriptor, we then introduce a global pooling layer to the network. Each sampled window is separately fed to the proposed network for feature extraction and then aggregated at the global pooling layer as shown in Fig \ref{fig:model}. Hence, the pooling layer has an important impact on the feature aggregation as well as the final prediction. There are two commonly used pooling strategies, max-pooling and average-pooling. In the proposed task, our preliminary tests showed that average-pooling outperforms max-pooling.  We believe that max pooling results in too great a loss of information; eliminating outliers is less important in this application, making max pooling less appealing.
Thus, we applied the global average-pooling for the feature aggregation, and we have,

\begin{equation}
H_G = \mbox{mean}(H^{(1)}, H^{(2)}, ... H^{(t)}, ..., H^{(K)}),
\end{equation}
where K is the number of windows from the traces. Then we feed the feature vector into a full-connected layer and a softmax layer. And finally we obtain the softmax cross-entropy loss
\begin{equation}
L = -\sum_{i}y_{i}log(s_i)
\end{equation}

\section{Results}
This section presents our experimental results on team performance prediction using the proposed ST-GCN model.  We compare our proposed model to standard benchmarks and ablated versions of the neural architecture.

\subsection{Dataset}
We evaluate the proposed model on two search and rescue missions (Mission A and B) conducted in the Minecraft team experimentation environment built by Aptima \cite{ASU/BZUZDE_2022}. For each mission, three participants must cooperate to rescue victims hidden in a large building; the subjects only have partial knowledge of the map in advance. The team scores 10 points by rescuing a normal victim and 50 points by cooperating to rescue a critically wounded victim.

 The dataset includes data from 60 teams performing both the missions in randomized order. Each team has 15 minutes to complete the SAR task; player trajectory and FOV (field of view) information is recorded as the participants move.  The trajectory information includes the coordinates of the participants at every single timestamp. Based on this, we can estimate the velocity of the players. Thus we obtain the trajectory feature $X^{(t)}_{traj} = \left\{x, y, v, v_x, v_y \right\}$ as mentioned in Sec.~\ref{sec:gcn}. We also applied the PyGL-FoV package to extract features describing the participant's field of view (FoV).
 While a participant moves, the PyGL-FoV agent generates messages of the blocks of interest that appear in the player's viewport, as shown in Fig \ref{fig:feature}. We extract the number of victims that appear in the viewport to include in the node feature of the graph. Thus, for each node, we have the feature vector $X^{(t)} = X^{(t)}_{fov} + X^{(t)}_{traj}$, with size of $6\times 1$. The PyGL-FoV agent is available in Python and rendered using OpenGL v1.5.

We divided the data files into 30 second segments. Since the interval between each window is 2 seconds, we have 15 windows in each sample. The performance of the team is then identified as high-performance or low-performance based on the points gained; this enables us to monitor team performance at a sufficiently fine granularity to empower an agent to assist poorly performing teams. The threshold in our experiments is set as 10 points. We have 1619 and 1672 samples for mission A and mission B respectively. All datasets are randomly partitioned into training set and test set by $4:1$ ratio, with no overlap between them. All of the results described in this section were obtained on the test set.

\subsection{Implementation Details}

Our network takes 15 windows as the input. For each window, we have 3 participants, thus the dimensions for the input feature matrix of the GCN is $15 \times 3 \times 6$. Each feature matrix is fed into a ST-GCN module, with 32 hidden units. Thus, the feature dimension for each window is $32 \times 3$. To aggregate the features from each window into a compact vector for prediction over the the entire trace sample, we applied global average pooling over all the flattened window features. 

Our model is trained for 1000 iterations in total, with the learning rate 0.0001. The Adam optimizer is used to train the proposed ST-GCN model. The batch size during training is set as 64. The model is implemented with the PyTorch framework and PyTorch Geometric \cite{rozemberczki2021pytorch}. We have made our code publicly available at: \href{https://github.com/shengnanh20/ST-GCN-ASIST.git}{https://github.com/shengnanh20/ST-GCN-ASIST.git}).

All the experiments are evaluated based on three commonly used metrics: (1) Accuracy (Acc); (2) Root Mean Squared Error (RMSE); (3) F1 score. Specifically, accuracy represents the percentage of correct predictions, and the F1 score is a harmonic mean of the precision and recall. The best value of these two metrics is 1, and 0 the worst. RMSE is used to measure the error of the models. The smaller the value is, the better the method performs.

\subsection{Performance Comparison}
To validate the effectiveness of the proposed model on teamwork prediction, we compare ST-GCN with the following benchmarks: (1) RF: Random Forest \cite{breiman2001random}; (2) SVC: Support Vector Classification \cite{svensen2007pattern}; (3) FNN: Feed-forward Neural Network with two hidden layers; (4) GCN: the original Graph Convolution Network \cite{kipf2016semi} with two hidden layers; (5) GRU: a Gated Recurrent Unit network \cite{cho2014}; (6) DCRNN: a spatial-temporal graph network which employs diffusion convolutions \cite{li2018diffusion}. Table ~\ref{tab:comp} shows the results of our comparison. It can be observed that the deep neural network based methods including FNN, GCN, DCRNN and ST-GCN, tend to have better performance than the linear baseline models including RF and SVC. Our proposed ST-GCN model outperforms all the other methods on the three validation metrics for the two missions. This occurs because the FNN has difficulty dealing with the temporal information and long-term data. The GCN method only considers the spatial features, which fail to capture the temporal dependency between time windows.  DCRNN is more effective than the previous methods at capturing temporal dependencies and spatial relations; however we believe that diffusion convolution is better suited for static graphs. Our proposed method outperforms DCRNN by about 3\% on both Mission A and Mission B.

\begin{table}[]
\centering
\caption{Performance comparison of different approaches.}
\label{tab:comp}
\begin{tabular}{l|ccl|ccl}
\hline
\multicolumn{1}{c|}{\multirow{2}{*}{Method}} & \multicolumn{3}{c|}{Mission A} & \multicolumn{3}{c}{Mission B} \\ \cline{2-7} 
\multicolumn{1}{c|}{}                        & Acc           & RMSE   &F1        & Acc           & RMSE  &F1         \\ \hline
RF                               & 0.59              &  0.62 &0.67              &0.55               & 0.67 &0.70              \\
SVC                                          &0.66 &0.58 &0.65               & 0.52               & 0.70              & 0.34              \\
FNN             & 0.72              & 0.52  &0.72              & 0.70   &0.55    &0.71      \\

  GCN   &0.69   &0.55     &0.69        &0.69  & 0.56      &  0.72                                                  \\
 GRU                             &0.71               & 0.53                & 0.74               & 0.69  &0.56 &0.70        \\
 DCRNN  & 0.72   & 0.53    & 0.70  & 0.71 &0.54 &0.71                   \\
ST-GCN                                       & \textbf{0.75}             & \textbf{0.49}               &  \textbf{0.75}         & \textbf{0.74} & \textbf{0.51} & \textbf{0.77}                  \\ \hline
\end{tabular}
\end{table}

\subsection{Effect of the GCN Module}
To investigate the effect of the graph convolution, we construct a model replacing the layers of GCN with fully-connected layers and then train and test the model under the same experimental settings.
As shown in Table \ref{tab:comp}, the performance of the proposed ST-GCN model is obviously higher than the model without the GCN module (GRU only). This is unsurprising because the linear layer and GRU module both experience difficulties modeling the spatial dependencies between participants.

\subsection{Effect of the Temporal GRU Module}
To further evaluate the effect of the GRU module in the proposed method, we trained the proposed model without the GRU for comparison. Specifically, we remove the GRU module and extract features directly with the GCN baseline model. As shown in Table \ref{tab:comp}, compared to the ST-GCN without GRU module, the proposed method reaches higher accuracy and F1 score, with a lower RMSE. The intuition is that the GCN only considers the spatial dependence, which fails to incorporate the temporal information between windows. This further indicates the effectiveness of the proposed method on spatial and temporal feature extraction.

\subsection{Parameter Analysis}

\subsubsection{The Impact of Input Length}
\begin{figure}[]
\subfigure[ROC on Mission A]{
	\begin{minipage}[t]{0.5\textwidth}
		\centering
		\includegraphics[scale=0.3]{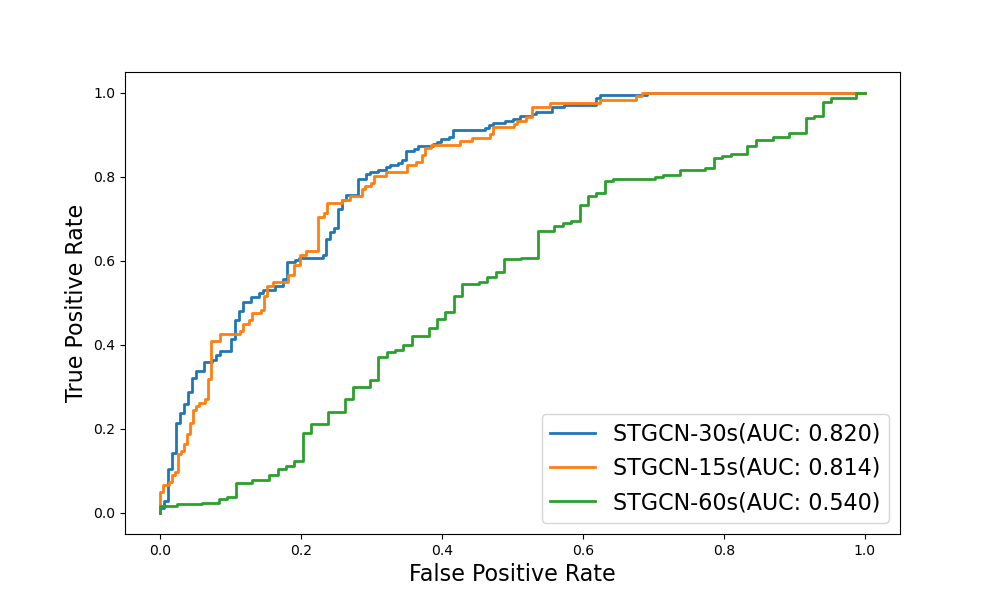}
	\end{minipage}
	}
	\newline
\subfigure[ROC on Mission B]{
	\begin{minipage}[t]{0.5\textwidth}
		\centering
		\includegraphics[scale=0.3]{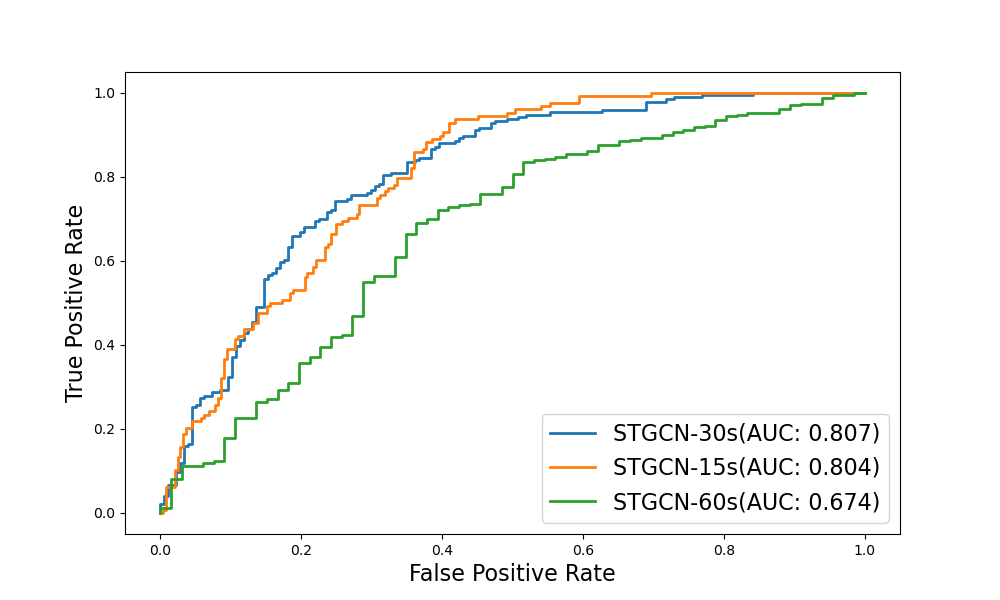}
	\end{minipage}}
\centering
\caption{ROC curves of ST-GCN trained with different input lengths.}
\label{fig:time}
\end{figure}

To study the influence of time horizon on performance prediction, experiments are carried out with different segment lengths.  Specifically, we set the interval as 1 second, 2 seconds and 4 seconds respectively to obtain the features within the segment length of 15 seconds, 30 seconds and 60 seconds. Training and testing are both conducted on the data with the same segment length. Note that the parameters of the ST-GCN are shared across all the experiments. The comparison results are presented in Table \ref{tab:time} and the ROCs are shown in Figure~\ref{fig:time}. It can be observed that the proposed ST-GCN with a segment length of 30 seconds performs the best on both the validation metrics and the ROC curves. The intuition is that when the segment length is 15 seconds, it is too short to offer sufficient information for the model. On the other hand, a longer time horizon can be detrimental, since more outliers may occur and there is more redundant information.



\begin{table}[]
\centering
\caption{Performance comparison of input lengths.}
\label{tab:time}
\begin{tabular}{l|ccl|ccl}
\hline
\multicolumn{1}{c|}{\multirow{2}{*}{Input Length}} & \multicolumn{3}{c|}{Mission A} & \multicolumn{3}{c}{Mission B} \\ \cline{2-7} 
\multicolumn{1}{c|}{}                              & Acc           & RMSE  &F1          & Acc           & RMSE  &F1          \\ \hline
15s                                                & 0.73       &0.51   &0.60    &  0.71              &0.54               &0.61               \\
30s                                                &  \textbf{0.75}   &\textbf{0.49}  &\textbf{0.75}            & \textbf{0.74} &\textbf{0.51} &\textbf{0.77}             \\
60s                                                &  0.53             &    0.69 &0.69            & 0.57    &0.66          &0.71               \\
\hline
\end{tabular}
\end{table}


\subsubsection{The Impact of the Input Feature}
\begin{figure}[]
\subfigure[ROC on Mission A]{
	\begin{minipage}[t]{0.5\textwidth}
		\centering
		\includegraphics[scale=0.3]{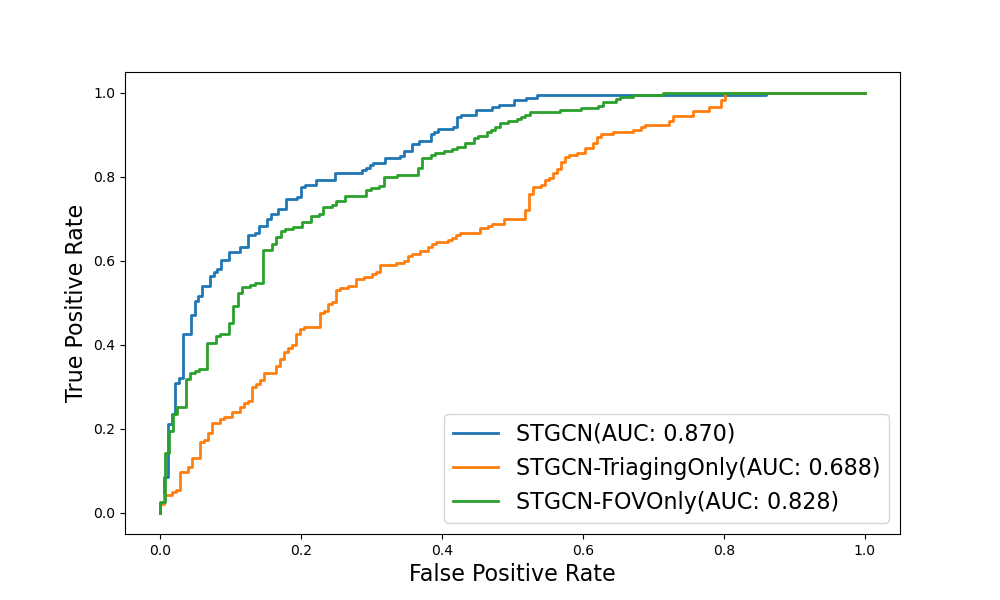}
	\end{minipage}
	}
\subfigure[ROC on Mission B]{
	\begin{minipage}[t]{0.5\textwidth}
		\centering
		\includegraphics[scale=0.3]{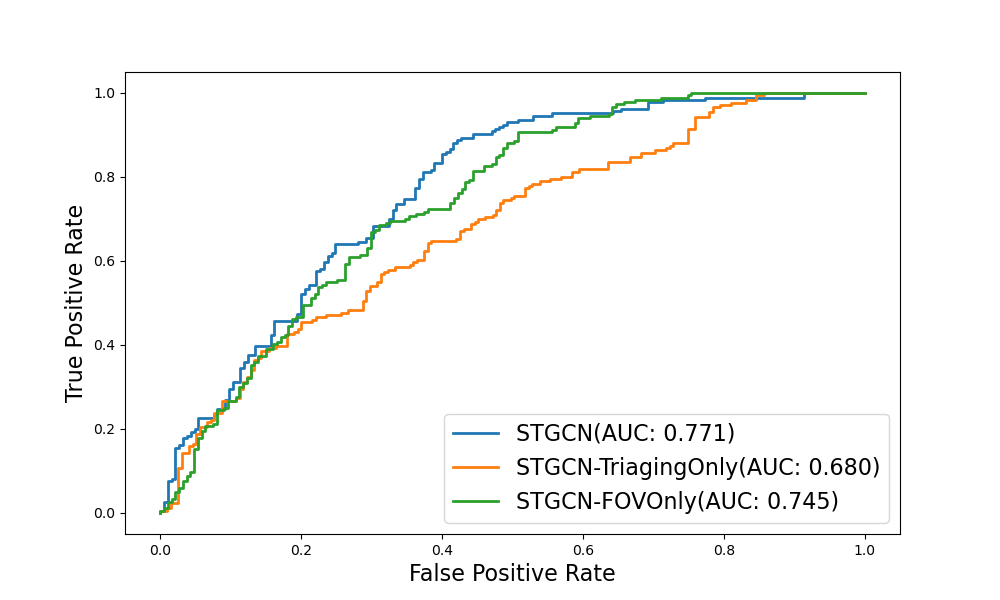}
	\end{minipage}}
\caption{ROC curves of ST-GCN trained with different input features.}
\label{fig:feat}
\end{figure}

To verify the effect of the input features on prediction, we compare the performance of our proposed ST-GCN trained with different feature vectors. The results are given in Table \ref{tab:feature} and Fig. \ref{fig:feat}. It is observed that FOV features outperform the triaging features. This is reasonable because compared with triaging features, the FOV of the participants provides more explicit information about which victims have been detected than the triaging trajectories.  However, from Table \ref{tab:feature} and Fig. \ref{fig:feat}, we can see that the proposed model trained with both triaging and FOV features outperforms the model with only the FOV feature, which confirms that there is value in retaining the triaging information.  Several of our pilot studies have shown that player velocity can be correlated with high Minecraft performance so it is unsurprising that the velocity component of the triaging feature is valuable.

\begin{table}[]
\centering
\caption{Performance comparison of different input features.}
\label{tab:feature}
{\scriptsize
\begin{tabular}{l|ccl|ccl}
\hline
\multicolumn{1}{c|}{\multirow{2}{*}{Input Feature}} & \multicolumn{3}{c|}{Mission A} & \multicolumn{3}{c}{Mission B} \\ \cline{2-7} 
\multicolumn{1}{c|}{}                               & Acc           & RMSE  & F1         & Acc           & RMSE  & F1        \\ \hline
Triaging Features Only                               & 0.63               & 0.61 &0.68                &0.60         &0.63      &  0.64             \\
FOV Features Only   & 0.67  &0.58  &0.67              & 0.68    &0.56          &0.69               \\
Triaging Feature + FOV Feature                       & \textbf{0.75}             & \textbf{0.49}               &  \textbf{0.75}         & \textbf{0.74} & \textbf{0.51} & \textbf{0.77}                      \\ \hline
\end{tabular}
}
\end{table}

\section{Conclusion and Future Work}
This paper introduces a new approach for predicting team performance of players performing a joint search and rescue task in Minecraft. We show that ST-GCN (Spatial Temporal Graph Convolutional Network) is capable of learning complex spatial, temporal, and coordination relationships from modest amounts of data.  The GCN is used to aggregate multimodal feature data using an adjacency matrix inversely proportional to player separation.   Thus when players are closer together their features assume a higher weight.  Our model can infer player performance in real-time, making it a suitable choice for a proactive assistant agent.  Although increasing the time segment length is problematic, ST-GCN could be combined with a time series predictor such as ARIMA to make predictions at a longer time horizon. We demonstrate our model on small teams with three players, but we believe that the GCN would easily scale to modeling larger coalitions commonly found in massively multiplayer online games. 

 Previous work on this dataset has focused on the problem of inferring player beliefs~\cite{falsebelief} and analyzing team communication~\cite{emergentleadership} towards the creation of agents that possess computational theory of mind and artificial social intelligence, whereas our framework does not currently incorporate features about player roles and communication.  We believe that the ST-GCN could easily be extended to leverage these features by augmenting the relationships encoded in the GCN graph.

\section{Acknowledgement}
This material is based upon work supported by the Defense Advanced Research Projects Agency (DARPA) under Contract No. W911NF-20-1-0008. 

\bibliographystyle{IEEEtran}
\bibliography{ref}



\end{document}


%
\title{Predicting Team Performance with Spatial Temporal Graph Convolutional Networks 
\\(Supplementary Material)
}

\author{\IEEEauthorblockN{Shengnan Hu}
\IEEEauthorblockA{Department of Computer Science\\
University of Central Florida\\ 
Orlando, FL, USA\\
Email: shengnanhu@knights.ucf.edu}
\and
\IEEEauthorblockN{Gita Sukthankar}
\IEEEauthorblockA{Department of Computer Science\\
University of Central Florida\\ 
Orlando, FL, USA\\
Email: gitars@eecs.ucf.edu}}


%


\maketitle


\subsection{Dataset}
The dataset used in our paper was collected from search and rescue missions executed in a custom Minecraft experimentation environment created by Aptima~\cite{ASU/BZUZDE_2022,asisttestbed}.  This section provides more details on the data collection process than we were able to include in our main paper.

\begin{figure}[htbp]
\subfigure[Map of Mission A]{
	\begin{minipage}[t]{0.45\textwidth}
		\centering
		\includegraphics[scale=0.45]{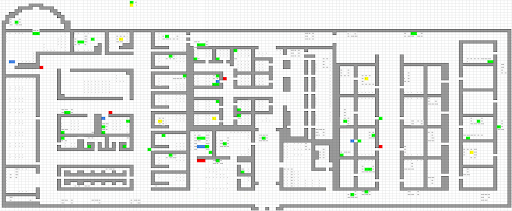}
	\end{minipage}
	}
	\newline
\subfigure[Map of Mission B]{
	\begin{minipage}[t]{0.45\textwidth}
		\centering
		\includegraphics[scale=0.45]{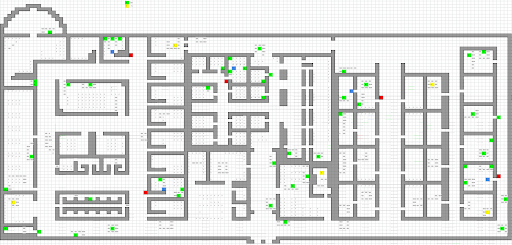}
	\end{minipage}}
\centering
\caption{Maps annotated with victim locations. Figure courtesy of Aptima \cite{ASU/BZUZDE_2022}.}
\label{fig:map}
\end{figure}

\subsubsection{Trajectory Data}
Experiments were conducted on the same map but with different victim placements (Mission A and Mission B); the victim configurations are shown in the Fig. \ref{fig:map}. Specifically, in the map, green blocks denote the regularly wounded victims, and yellow blocks denote critical victims that need to be rescued rapidly before they expire.  A successful team should triage yellow victims first, while marking the location of green victims for later rescue. In total, there are 50 regular victims and 5 critical victims in each mission. The gray blocks denote impassable regions.

Each team consists of three human subjects recruited to participate in a two hour experiment that includes intake and exit surveys, task instruction, and a Minecraft competency test.  During the fifteen minute mission, the team is asked to explore the map and rescue the victims.  There are three different team roles: medic (heals victims), searcher (has faster movement to empower faster exploration), and engineer (breaks obstacles).  The testbed records information about the participants' locations and actions.  After a team completes the experiment, the event data is stored into files on Google Cloud.  From these files, we extracted the location and velocity of the players to create the triaging feature vectors used by our proposed framework, ST-GCN.

\subsubsection{FOV data}
\begin{figure}
\centering
\includegraphics[width=0.45\textwidth]{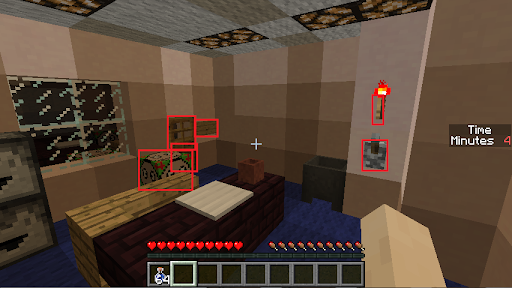}
\caption{Example player viewport marked by CMU's PyGL-FoV agent.}
\label{fig:fov}
\end{figure}

To obtain the victim information for FOV node features, we utilize the PyGL-FoV agent developed by CMU to extract the Field of View information; the code is available at: \url{https://gitlab.com/cmu_asist/PyGLFoVAgent}. PyGL-FoV generates a summary of Minecraft blocks that appear in the particpants viewport. To calculate the block list, PyGL-FoV renders the blocks in the Minecraft world using OpenGL and extracts information from the rendered scene by mapping the rendered pixels back to the Minecraft blocks. An example player viewport annotated by the PyGL-FoV agent is shown in Fig. \ref{fig:fov}.

\bibliographystyle{IEEEtran}
\bibliography{ref}


